%% file: main.tex
\documentclass{article}
\usepackage[final,sglblindworkshop]{neurips_2026}
\workshoptitle{AIDaR: AI Data Readiness for Scientific Discovery}
\makeatletter
\renewcommand{\@trackname}{Accepted at AI Data Readiness for Scientific Discovery Workshop \& \@neuripsordinal\ Conference on Neural Information Processing Systems (NeurIPS 2026), Paris.}
\makeatother
\usepackage{natbib}
\usepackage{pgfplots}
\pgfplotsset{compat=1.18}
\usepackage{listings}
\usepackage{placeins}
\usepackage{todonotes}
\usepackage{amsmath, amssymb, amsthm}
\usepackage{booktabs}
\usepackage{array}
\usepackage{fancyvrb}
\usepackage{hyperref}
\usepackage{amsmath, amssymb}
\usepackage{algorithm}
\usepackage{algpseudocode}
\usepackage{booktabs}
\usepackage{parskip}
\usepackage{longtable}
\usepackage{hyperref}
\usepackage{listings}
\algrenewcommand\algorithmicindent{1.2em}

\algnewcommand\algorithmicforeach{\textbf{for each}}
\algdef{S}[FOR]{ForEach}[1]{\algorithmicforeach\ #1\ \textbf{do}}
\usepackage{parskip}
\usepackage{enumitem}
\usepackage{xcolor}
\usepackage{stmaryrd}
\usepackage{mdframed}
\usepackage{tikz}
\usetikzlibrary{shapes, arrows.meta, positioning}
\usepackage{booktabs}
\usepackage{tabularx}
\usepackage[table]{xcolor}
\usepackage{float}
\newcolumntype{L}[1]{>{\raggedright\arraybackslash}p{#1}}
\newtheorem{definition}{Definition}

\definecolor{correct}{RGB}{0,120,0}
\definecolor{wrong}{RGB}{180,0,0}
\definecolor{reasoner}{RGB}{0,0,160}

\definecolor{hilight}{RGB}{230,245,230}
\definecolor{correct}{RGB}{0,120,0}
\definecolor{wrong}{RGB}{180,0,0}

\newcolumntype{C}[1]{>{\centering\arraybackslash}p{#1}}
\begin{document}

\title{Ontology-Grounded, Reasoner-Verified Benchmarks for Evaluating LLM Reasoning in Scientific AI}

\author{%
  Nishtha N. Vaidya$^{1,2}$, \; Stephan Grimm$^{1}$, \; Thomas Hubauer$^{1}$, \; Thomas A. Runkler$^{1,2}$ \\[2pt]
  $^{1}$Siemens AG \quad
  $^{2}$Technical University of Munich \\
  Email: \texttt{nishtha.vaidya@tum.de}
}

\maketitle
\begin{abstract}
Large language models (LLMs) increasingly underpin scientific AI applications that reason over structured knowledge, from biomedical question answering and clinical decision support to materials informatics. However, their logical reasoning often falls short, producing factual inaccuracies unacceptable in these settings. Reliable evaluation remains challenging: manual dataset construction scales poorly, and LLM-based generation risks embedding the very flaws it aims to measure. High-quality benchmarks must therefore ground both correct and incorrect labelled examples in explicit background knowledge, formally verifiable by a standard reasoner.
We propose a pipeline that automatically generates ontology-grounded multiple-choice question (MCQ) benchmarks from any sufficiently axiomatised OWL 2 ontology, with correct answers grounded in the ontology by design. Incorrect answers, called distractors, are generated by perturbing the right-hand-side class expressions of class definition axioms, and their incorrectness with respect to the ontology's axioms is formally verified by an OWL reasoner via entailment checks.
We evaluate the pipeline on three ontologies: Pizza (small, academic), PMDco (complex, materials science), and DOID (large, biomedical). The results validate three claims.
First, the pipeline is productive and grounded: we use it to generate 112, 2,491, and 15,216 MCQs respectively, each anchored to a class definition axiom. Second, distractors span four semantic categories from class unsatisfiability to weakened subsumptions, enabling diagnostic evaluation of specific reasoning failures. Third, items meet natural language quality standards: mean LLM judge scores of 4.02, 4.36, and 3.36 out of 5 confirm fluency, and correct-answer-to-distractor similarity above 0.8 shows that wrong options cannot be dismissed on surface form alone.
Six LLMs (frontier, open-source, reasoning, and non-reasoning) evaluated zero-shot achieve 41.1--76.8\% accuracy, well above the 25\% random-guessing baseline, indicating the benchmarks are challenging and discriminative. This work is a  step towards more reliable benchmarks for assessing logical reasoning in next generation scientific AI applications.
\end{abstract}

\input{introduction}

\input{readme}

\input{method}

\input{experiments}

\input{conclusions}

\bibliographystyle{plainnat}
\bibliography{articles}

\appendix
\input{appendix}
\end{document}

%% file: introduction.tex
\section{Introduction}
Large language models (LLMs) perform well on natural language (NL) tasks but fail on logical reasoning over domain knowledge \citep{liu2025logical}, producing factual inaccuracies unacceptable for scientific applications such as biomedical knowledge retrieval, clinical decision support, and materials informatics.
This holds even when models are augmented with prompt engineering \citep{10970773}, retrieval-augmented generation \citep{toro2024dynamic}, or multi-step reasoning pipelines \citep{singh2025agentic}.
Structured domain ontologies, widely used across biomedicine, materials science, and other scientific fields, encode the formal knowledge underpinning scientific discovery workflows such as entity classification and data annotation, making correct LLM reasoning over these ontologies a prerequisite for reliable scientific AI.
 
Benchmarking LLM logical reasoning over domain ontologies requires datasets that: 1) ground entries in explicit background knowledge, and 2) express statements in formal logic verifiable by a reasoner.
Such datasets serve multiple communities: scientific AI developers evaluating model fitness for scientific reasoning tasks; model builders extracting training signals to penalise specific reasoning errors; domain scientists validating LLM outputs against their knowledge base without formal logic expertise; and neuro-symbolic researchers evaluating hybrid architectures \citep{vsevolodovna2025enhancing}.
Manual benchmark construction requires expert verification of logical soundness, which is labour-intensive and does not scale. LLM-based auto-generation risks embedding the same reasoning flaws the benchmark aims to measure, since LLMs are prone to hallucinations.

We propose a pipeline that automatically generates multiple-choice question (MCQ) benchmarks from any sufficiently axiomatised Web Ontology Language (OWL 2) ontology. Each MCQ has precisely one answer that is correct with respect to the ontology's axioms, and distractors verified as incorrect against those same axioms by a reasoner. Each pair is provided in two complementary forms: an NL verbalisation and the underlying symbolic representation.
The correct answers are derived from ontology class definition axioms, verbalizing their concept expression structure. Distractors are generated by perturbing the right-hand-side class expressions of these axioms and their incorrectness is formally verified by an OWL reasoner via entailment checks. Distractors span four semantic categories, from class unsatisfiability to weakened subsumptions, enabling diagnostic evaluation of specific reasoning failures.
We evaluate the pipeline on three ontologies: Pizza (academic) \citep{horridge2009practical}, PMDco (materials science) \citep{https://doi.org/10.1002/adem.71106}, and the Disease Ontology (DOID, biomedical) \citep{schriml2022human}. The pipeline is domain-agnostic: applied to any sufficiently axiomatised OWL 2 ontology, it produces a benchmark grounded in that domain's scientific knowledge. Named domain concepts are retained by design, reflecting the target deployment context: scientific AI systems reason over specific ontologies where domain concepts are named, and a benchmark that strips them would not reflect this setting. This work offers a significant step towards developing more reliable and accurate
benchmarks for assessing the logical reasoning capabilities of next-generation scientific AI systems.
The paper is organised as follows: Section 2 reviews the related work. Section 3 presents the proposed pipeline. Section 4 and Section 5 report the experiments and the conclusions.

%% file: readme.tex
\section{Related Works}

Existing benchmarks fail at least one of three requirements: grounding in domain ontologies, formal verification of both correct answers and distractors, and scalable automated construction.
OntoLAMA \citep{he2023language} probes subsumption entailment as a binary classification task, deriving positive samples with a symbolic reasoner.
Its negative samples rely on assumed disjointness theory but are not formally verified in practice.
LLM-ORBench \citep{loesch2026llmorbench} grounds true/false and open-ended questions in ontology-derived triples verified by a reasoner.
It does not specify how false examples are constructed, leaving distractor type undefined.
OntoURL \citep{zhang2025ontourl} broadens coverage to 57,303 MCQs and applies a reasoner to derive inferred relations.
Its distractors are drawn from structurally related entities such as ancestors and siblings, without formal verification of their incorrectness. It uses named classes as distractors, reducing the task to named entity recall.
MALLS \citep{yang2024harnessing} generates NL--first-order logic (FOL) pairs via LLM; its authors acknowledge these as silver-label data requiring gold-standard benchmarks for reliable evaluation.
FOLIO \citep{han-etal-2024-folio} provides formally grounded, verifier-checked natural language reasoning examples.
However, its 1,435 examples required approximately 980 expert annotation hours, and it targets general FOL over Wikipedia rather than structured domain knowledge.

%% file: method.tex
\section{A Method for Generating MCQs from Ontology Class Definition Axioms}

We present a four-stage pipeline to automatically generate an MCQ benchmark from any sufficiently axiomatised OWL 2 ontology.
The high-level workflow (Figure~\ref{fig:workflow}, top) proceeds as follows:
\begin{description}

    \item[\textbf{Stage 1}] \textbf{(Axiom extraction and perturbation):} Class definition
    axioms are selected from the ontology to ground MCQ generation. Each axiom is
    perturbed to generate concept distractors.

    \item[\textbf{Stage 2}] \textbf{(Verbalisation and smoothing):} The description logic (DL) expressions
    for both axioms and distractors are verbalised into fluent NL.

    \item[\textbf{Stage 3}] \textbf{(Distractor ranking and selection):} Not all
    distractors are equally challenging, so they are ranked to identify the top-$k$
    hardest-to-reject distractors.
\begin{figure}[!htp]
    \centering
    \includegraphics[width=1\linewidth]{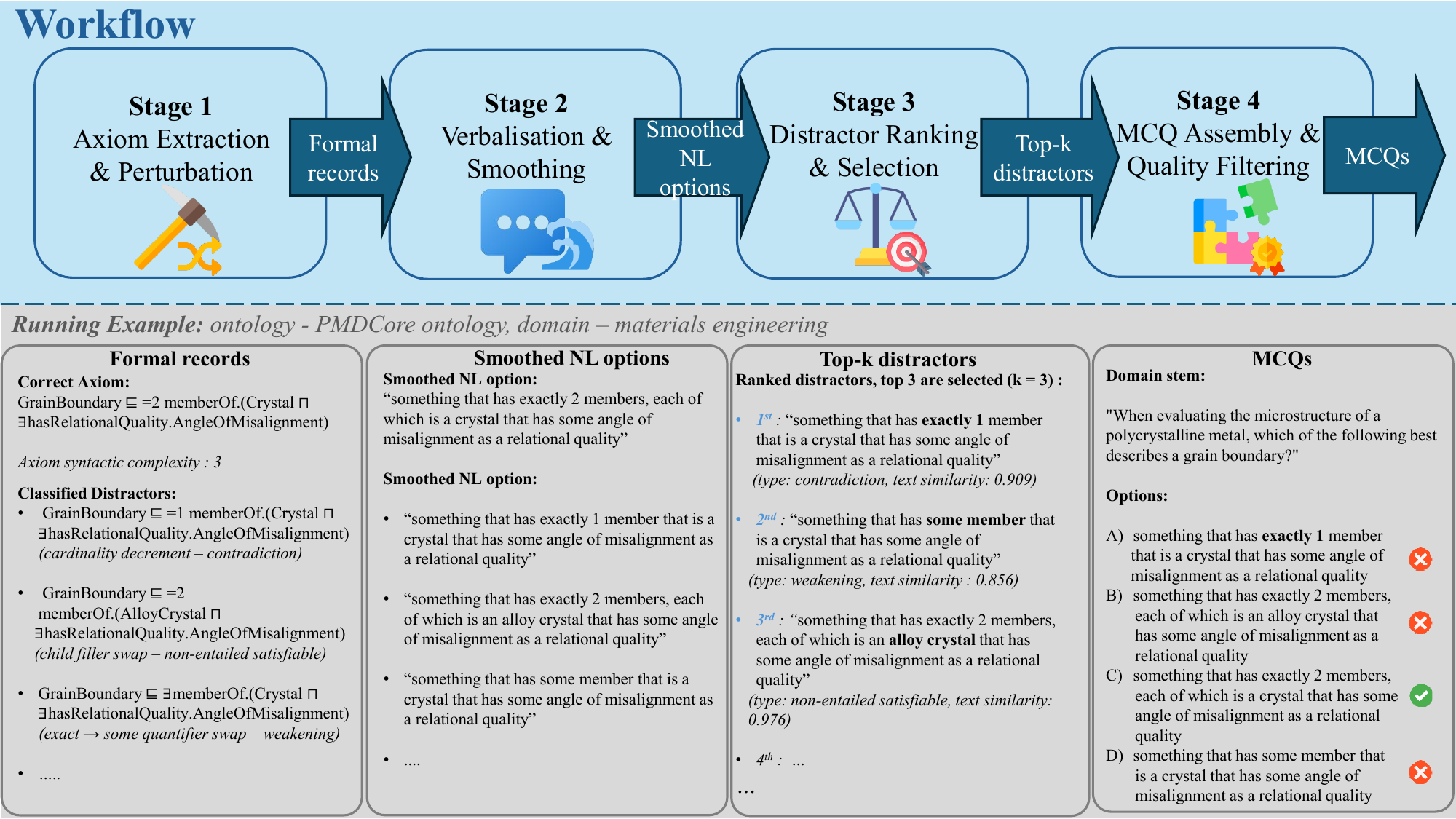}
    \vspace{-1.5em}
    \caption{Four-stage pipeline for generating ontology-grounded 
multiple-choice questions. Stages proceed left to right and the bottom row 
illustrates each stage with an example from
PMDco.}
    \label{fig:workflow}
\end{figure}
    \item[\textbf{Stage 4}] \textbf{(MCQ assembly and quality filtering):} A question
    stem (the instruction that precedes the options and defines what is being evaluated)
    is generated for each record. Options are shuffled and assembled into a complete MCQ;
    each is scored by an LLM judge and only high-quality items are retained.

\end{description}

The MCQ format makes evaluation objective and reproducible: exactly one correct answer, no additional annotation, and difficulty controllable through distractor design. Existing ontology-based benchmarks share this format but use named classes as distractors; our pipeline instead generates complex class expressions, shifting the challenge from named entity recall to ontological reasoning. The following subsections describe each stage, illustrated by a running example from PMDco (Figure~\ref{fig:workflow}, bottom row).

\subsection{Generating Concept Distractors by Perturbation}
\label{sec:distractorperturbation}

We consider class definition axioms of the form $A \sqsubseteq C$, where $A$ is a named
OWL class and $C$ is a complex class expression.
We assume access to the inferred class hierarchy of named classes in $\mathcal{O}$:
$\mathcal{H} = \{ A_1 \sqsubseteq A_2 : \mathcal{O} \models A_1 \sqsubseteq A_2,\ A_1, A_2 \in \sigma(\mathcal{O})\}$.

For example, consider the class definition axiom from the PMDco materials ontology:
\begin{equation}
    \textit{GrainBoundary} \sqsubseteq\ {=}2\, \textit{memberOf}.(\textit{Crystal}\ \sqcap\
    \exists\textit{hasRelationalQuality}.\textit{AngleOfMisalignment})
    \label{eqn:1}
\end{equation}
This axiom defines a grain boundary as a class with exactly two members, each a crystal bearing some angle of misalignment as a relational quality.
$\mathcal{H}$ provides, for example, that every alloy crystal is also a crystal
($\textit{AlloyCrystal} \sqsubseteq \textit{Crystal}$).

We generate distractor candidates for a class definition axiom $A \sqsubseteq C$ by applying perturbation operators to $C$.
\begin{definition}[Concept Distractor and Perturbation Operator]
A \emph{concept distractor} for $C$ is a concept expression $C'$ in the vocabulary of $\mathcal{O}$ with $C \not\equiv C'$.
A \emph{perturbation operator} $\pi$ maps $C$ to a finite set $\pi(C) = \{C'_1, C'_2, \ldots\}$, where each $C'_i$ is expressed in the vocabulary of $\mathcal{O}$.
We write $C[p \leftarrow v']$ for replacing the value $v$ at position $p$ with $v'$ (single-position substitution), and $C[v \twoheadleftarrow v']$ for replacing all occurrences of $v$ in $C$ (global substitution).
\label{def:distractor}
\end{definition}

For example, the concept expression in Eq.~\ref{eqn:dist} is a concept distractor for the right-hand side of Eq.~\ref{eqn:1}: it uses only vocabulary of $\mathcal{O}$ but requires \emph{some} crystal member rather than \emph{exactly two}.
\begin{equation}
    \exists\,\textit{memberOf}.(\textit{Crystal}
\sqcap \exists\,\textit{hasRelationalQuality}.\textit{AngleOfMisalignment})
\label{eqn:dist}
\end{equation}

Each operator $\pi$ targets a position $p$ in $C$, which may be a named class, role, cardinality bound, restriction kind (existential, universal, exact, min, or max), or Boolean constructor, and draws a replacement $v'$ from an operator-specific source set $\mathcal{S}_\pi(p, C)$, yielding $C'_i = C[p \leftarrow v']$ or $C[v \twoheadleftarrow v']$.
Each operator is characterised by its target type, source set $\mathcal{S}_\pi$, an optional guard condition on $C$ or $p$, and an optional output cap that bounds $|\pi(C)|$; Table~\ref{tab:operators} summarises all operators.

Consider operator $\pi_{Q_1}$ applied to Eq.~\ref{eqn:1}.
The target position $p$ is the exact cardinality constructor ${=}2$ in ${=}2\;\textit{memberOf}$; the source set $\mathcal{S}_{\pi_{Q_1}} = \{\exists\}$, so the replacement $v'$ is the existential constructor $\exists$.
The substitution $C[p \leftarrow v']$ yields Eq.~\ref{eqn:dist}, leaving the rest of the expression unchanged.
\begin{table}[!tp]
\scriptsize
\centering
\caption{Formal Perturbation Operators $\pi$.
 }
\label{tab:operators}
\renewcommand{\arraystretch}{1.4}
\begin{tabularx}{\textwidth}{@{} l >{\hsize=1.3\hsize\linewidth=\hsize}X >{\hsize=1.0\hsize\linewidth=\hsize}X >{\hsize=0.7\hsize\linewidth=\hsize}X @{}}
\toprule
\textbf{Op} & \textbf{Transformation $C'_i$} 
            & \textbf{Source Set $\mathcal{S}_\pi(p,C)$} 
            & \textbf{Guard / Cap} \\
\midrule

$\pi_{F_1}$ 
& Per distinct filler $B$ at position $p$: 
  $C[B \twoheadleftarrow v']$ for each $v' \in \mathcal{S}_\pi(p,C)$
& Siblings of $B$ and their descendants to depth $N$ in $\mathcal{H}$
& Skip if $C$ contains $\forall$; depth cap $= N$ \\

$\pi_{F_2}$ 
& Per distinct filler $B$ at position $p$: 
  $C[B \twoheadleftarrow v']$ for each $v' \in \mathcal{S}_\pi(p,C)$
& All ancestors of $B$ in $\mathcal{H}$; all levels
& None \\

$\pi_{F_3}$ 
& Per distinct filler $B$ at position $p$: 
  $C[B \twoheadleftarrow v']$ for each $v' \in \mathcal{S}_\pi(p,C)$
& Direct children of $B$ in $\mathcal{H}$
& Cap $= P$ children per filler
\\

$\pi_{F_4}$ 
& Per distinct filler $B$ at position $p$: 
  $C[B \twoheadleftarrow v']$ for each $v' \in \mathcal{S}_\pi(p,C)$
& Classes disjoint with $B$ in $\mathcal{H} : \mathcal{D}[B]$ 
& Skip if $C$ contains $\forall$; cap $= M$ per filler \\

\midrule

$\pi_{S_1}$ 
& $p$ = cardinality bound $n$ in top-level ${\bowtie}_n\, r.E$:
  $C[p \leftarrow n+1]$ and $C[p \leftarrow n-1]$
& $\mathcal{S}_\pi(p,C) = \{n-1,\, n+1\}$; 
  omit $n-1$ if $n-1 < 0$
& Top-level kind $\in \{\text{exact},\, \text{min},\, \text{max}\}$ \\

\midrule

$\pi_{Q_1}$ 
& $p$ = restriction kind in top-level ${=}_n\, r.E$: 
  $C[p \leftarrow \exists]$, dropping $n$
& $\mathcal{S}_\pi(p,C) = \{\exists\}$
& Top-level must be exact \\

$\pi_{Q_2}$ 
& $p$ = restriction kind in top-level ${\geq}_n\, r.E$: 
  $C[p \leftarrow \exists]$, dropping $n$
& $\mathcal{S}_\pi(p,C) = \{\exists\}$
& Top-level must be min \\

$\pi_{Q_3}$ 
& $p$ = restriction kind in top-level $\exists r.E$: 
  $C[p \leftarrow {=}]$, setting $n \leftarrow 1$
& $\mathcal{S}_\pi(p,C) = \{{=}1\}$
& Top-level must be $\exists r.E$ with $r$ non-transitive\\

\bottomrule
\end{tabularx}
\end{table}
We illustrate each operator on Eq.~\ref{eqn:1}.

\textit{Filler substitution ($\pi_{F_1}$ -- $\pi_{F_4}$).}
Operators $\pi_{F_1}$ through $\pi_{F_4}$ each target a named atomic class $B$
appearing as a filler in $C$ and replace it globally with a class drawn
from a position-specific source set in $\mathcal{H}$.
For example, $\pi_{F_3}$ targets the filler \textit{Crystal} and replaces it with each
of its direct children in $\mathcal{H}$.
Since $\mathcal{H}$ contains $\textit{AlloyCrystal} \sqsubseteq \textit{Crystal}$,
a candidate produced by $\pi_{F_3}$ is
\begin{equation}
    {=}2\;\textit{memberOf}.(\textit{AlloyCrystal} \;\sqcap\;
    \exists\,\textit{hasRelationalQuality}.\textit{AngleOfMisalignment})
    \label{eqn:f3}
\end{equation}
Only the filler is modified; the restriction kind, role, and remaining conjunct are unchanged.
$\pi_{F_1}$ replaces $B$ with the siblings of $B$ and their descendants.
$\pi_{F_2}$ replaces $B$ with each of its ancestors in $\mathcal{H}$.
By contrast, $\pi_{F_4}$ replaces $B$ with a class from the disjointness index
$\mathcal{D}[B]$, the set of classes disjoint with $B$ in $\mathcal{O}$.

$\pi_{F_1}$ and $\pi_{F_4}$ include a guard that skips any $C$ containing a universal
restriction $\forall r.E$.
In such expressions, replacing the filler with a sibling or disjoint class produces a
candidate rejectable from the expression structure alone, without ontology knowledge;
such distractors do not test the target reasoning skill and are excluded.
To keep the candidate pool tractable for large $|\mathcal{H}|$, the operators apply
output caps: descendants to depth $N$ for $\pi_{F_1}$, up to $P$ direct children for
$\pi_{F_3}$, and up to $M$ disjoint classes for $\pi_{F_4}$.

\textit{Cardinality bound modification ($\pi_{S_1}$).}
Operator $\pi_{S_1}$ targets the cardinality bound $n$ in a top-level restriction
${\bowtie}_n\, r.E$, where ${\bowtie} \in \{=, \geq, \leq\}$ is the restriction kind,
$r$ a role, and $E$ a concept expression.
It produces up to two candidates by replacing $n$ with $n+1$ and $n-1$.
Applied to the top-level restriction ${=}2\;\textit{memberOf}$ in $C$, $\pi_{S_1}$
produces
\begin{equation}
    C'_1 \equiv 
    {=}3\;\textit{memberOf}.(\textit{Crystal} \;\sqcap\;
    \exists\,\textit{hasRelationalQuality}.\textit{AngleOfMisalignment})
\end{equation}
\begin{equation}
    C'_2 \equiv {=}1\;\textit{memberOf}.(\textit{Crystal} \;\sqcap\;
    \exists\,\textit{hasRelationalQuality}.\textit{AngleOfMisalignment})
    \label{eq:s2}
\end{equation}
Both candidates are semantically distinct from $C$, since exact cardinality
restrictions with different bounds are mutually inconsistent; the $n-1$ variant is
skipped when $n = 0$ to avoid a negative bound.

\textit{Restriction kind replacement ($\pi_{Q_1}$--$\pi_{Q_3}$).}
Operators $\pi_{Q_1}$ through $\pi_{Q_3}$ each target the restriction kind at the
top-level restriction, leaving the role and filler unchanged.
$\pi_{Q_1}$ applies when the top-level restriction is an exact cardinality
${=}n\,r.E$ and replaces it with the existential $\exists r.E$, dropping the
cardinality bound.
Applied to $C$, $\pi_{Q_1}$ produces the single candidate in Eq.~\ref{eqn:dist}.
$\pi_{Q_2}$ applies analogously when the top-level restriction is a minimum
cardinality ${\geq}\,n\,r.E$, again replacing it with $\exists r.E$.
$\pi_{Q_3}$ operates in the reverse direction: when the top-level restriction is
already existential $\exists r.E$, it replaces it with the exact restriction
${=}1\,r.E$, introducing a cardinality constraint where none existed.
Since OWL~2 forbids cardinality restrictions on transitive properties, $\pi_{Q_3}$
skips existential restrictions whose role is transitive.

Each candidate $C'$ produced by a perturbation operator stands in a specific logical
relationship to the original axiom $A \sqsubseteq C$ under $\mathcal{O}$.
This relationship determines what kind of reasoning is required to reject $C'$ as a
wrong answer.
We formalise this as a semantic category
\begin{definition}\label{eqn:category}[Semantic Category of Candidate Distractors]
Let $\mathcal{O}$ be an ontology, $A \sqsubseteq C$ a class definition axiom, and $C'$
a concept expression obtained by applying a perturbation operator on $C$.
The \emph{semantic category} of $C'$ with respect to $\mathcal{O}$ and
$A \sqsubseteq C$ is
\begin{align*}
\text{category}(C') =
\begin{cases}
\textit{contradiction}
    & \text{if } \mathcal{O} \not\models A \sqsubseteq C'
      \text{ and } \mathcal{O} \models A \sqcap C' \sqsubseteq \bot \\[4pt]
\textit{non-entailed-satisfiable}
    & \text{if } \mathcal{O} \not\models A \sqsubseteq C'
      \text{ and } \mathcal{O} \not\models A \sqcap C' \sqsubseteq \bot \\[4pt]
\textit{weakening}
    & \text{if } \mathcal{O} \models A \sqsubseteq C'
      \text{ and } \mathcal{O} \models C \sqsubseteq C'
      \text{ and } \mathcal{O} \not\models C' \sqsubseteq C \\[4pt]
\textit{incomparable}
    & \text{if } \mathcal{O} \models A \sqsubseteq C'
      \text{ and } \mathcal{O} \not\models C \sqsubseteq C'
      \text{ and } \mathcal{O} \not\models C' \sqsubseteq C \\[4pt]
\textit{strengthening}
   & \text{if } \mathcal{O} \models A \sqsubseteq C'
      \text{ and } \mathcal{O} \not\models C \sqsubseteq C'
      \text{ and } \mathcal{O} \models C' \sqsubseteq C \\[4pt]
\textit{equivalent}
   & \text{if } \mathcal{O} \models A \sqsubseteq C'
      \text{ and } \mathcal{O} \models C \sqsubseteq C'
      \text{ and } \mathcal{O} \models C' \sqsubseteq C
\end{cases}
\end{align*}
\end{definition}
The category labels \emph{contradiction}, \emph{weakening}, and
\emph{strengthening}, and the entailment checks used to assign them, are borrowed
from ontology repair research \citep{troquard2018repairing,Colucci2003ConceptAA},
which characterises how a repair candidate relates to an inconsistent axiom.
We extend the taxonomy with \emph{non-entailed-satisfiable} and \emph{incomparable}, which complete the case analysis of the entailment check outcomes and have no counterpart in repair settings.

A distractor $C'$ is a \emph{contradiction} if $A$ and $C'$ are disjoint under
$\mathcal{O}$, making $A \sqcap C'$ unsatisfiable.
Operator $\pi_{S_1}$ produces such candidates: replacing the exact cardinality
bound $2$ with $1$ (Eq.~\ref{eq:s2}) means no individual can simultaneously
satisfy both the original and the perturbed definition.
By contrast, a distractor $C'$ is a \emph{weakening} if $C'$ is strictly more
general than $C$.
Operator $\pi_{Q_1}$ produces such candidates: replacing the exact cardinality
restriction with an existential (Eq.~\ref{eqn:dist}) yields a definition that
admits a broader range of individuals.
Not all distractors, however, stand in a clear-cut entailment or disjointness
relationship with $C$.
A distractor $C'$ is \emph{non-entailed-satisfiable} if $\mathcal{O}$ neither
entails $A \sqsubseteq C'$ nor derives a contradiction between $A$ and
$C'$: the perturbed expression is consistent with $\mathcal{O}$ but does not
hold for $A$.
This typically arises from sibling or child substitutions ($\pi_{F_1}$, $\pi_{F_3}$)
involving classes without disjointness axioms: replacing
\textit{Crystal} with a child class (Eq.~\ref{eqn:f3}) is consistent since no
disjointness axiom excludes it, but the ontology does not entail that every
grain boundary has exactly two alloy crystal members.
A distractor $C'$ is \emph{incomparable} if $\mathcal{O}$ entails
$A \sqsubseteq C'$ but $C$ and $C'$ are not related by subsumption in either
direction, describing unrelated aspects of $A$.
For example, if $\mathcal{O}$ also entails
$\textit{GrainBoundary} \sqsubseteq \exists\textit{partOf}.\textit{Polycrystal}$,
then $C' = \exists\textit{partOf}.\textit{Polycrystal}$ is incomparable with $C$.
A \emph{strengthening} makes $C'$ strictly more specific than $C$: for example,
adding $\exists\textit{partOf}.\textit{Polycrystal}$ as a conjunct
yields $C' = C \sqcap \exists\textit{partOf}.\textit{Polycrystal}$, which is
subsumed by $C$ but does not subsume it.
The most subtle case is \emph{equivalence}: $C'$ and $C$ subsume each other under
$\mathcal{O}$, making $C'$ semantically indistinguishable from the correct answer.

Each operator in Table~\ref{tab:operators} is applied to $C$, and each resulting candidate is evaluated against $\mathcal{O}$ using an OWL reasoner to determine its semantic category (Definition~3).
This requires four reasoner checks per candidate: $\mathcal{O} \models A \sqsubseteq C'$ (entailment), $\mathcal{O} \models A \sqcap C' \sqsubseteq \bot$ (satisfiability), and $\mathcal{O} \models C \sqsubseteq C'$ and $\mathcal{O} \models C' \sqsubseteq C$ (subsumption in both directions).
Candidates classified as \textit{equivalent} or \textit{strengthening} are also correct answers and are discarded; the remaining candidates, in the four valid categories, form the distractor pool for the axiom.\footnote{Operators that guarantee a fixed semantic category would remove the need for
post-hoc reasoning checks but must account for ontology entailments and may
still not yield sufficient candidates; we leave this to future work.}

\subsection{Verbalisation of Concept Expressions}
\label{sec:verbalise}
Both the correct expression and the distractors must be presented as readable text, so they are verbalised into NL as MCQ options.
Verbalisation begins with \textsc{DeepOnto} \citep{he2023language}, the established rule-based ontology verbaliser, which recursively applies fixed linguistic patterns to each constructor in the expression, producing deterministic output.
However, \textsc{DeepOnto} covers only the $\mathcal{ALC}$ fragment of OWL~2, so constructs such as \texttt{owl:hasSelf} and nominal sets may receive incomplete or unnatural verbalisations.
Each verbalisation is therefore \textit{smoothed} using an LLM with a fixed prompt template that receives both the \textsc{DeepOnto} verbalisation and the underlying DL expression; the LLM rewrites for fluency and grammatical completeness, with the logical content anchored by the expression rather than generated freely.\footnote{The LLM judge's option-faithfulness dimension (Section~\ref{sec:overall}) provides a quality signal for the smoothed verbalisations.}
For example, the distractor from Eq.~\ref{eqn:dist} is first verbalised as \emph{``something that member of crystal that has some relational quality angle of misalignment''} and then smoothed to \emph{``something that has some member that is a crystal that has some angle of misalignment as a relational quality''}.
\subsection{Metrics}\label{sec:metric}
We measure syntactic complexity by the \emph{operator count}
$\mathtt{oc}(C)$: the number of occurrences of the keywords
\texttt{some}, \texttt{only}, \texttt{and}, \texttt{or}, \texttt{not},
\texttt{min}, \texttt{max}, and \texttt{exactly}
in the Manchester Syntax rendering of $C$.
$\mathtt{oc}(C)$ is computable from the expression string alone
and is used to rank axioms by complexity and to match distractors
structurally to the correct expression (Section~\ref{sec:overall}).
All operators in Table~\ref{tab:operators} preserve the operator count by design: each
substitutes a value at a fixed structural position without adding or removing a
constructor, so every distractor has the same syntactic complexity as the correct
expression ($\mathtt{oc}(C') = \mathtt{oc}(C)$) and cannot be eliminated by comparing
structure alone.\footnote{$\mathtt{oc}(C)$ is a syntactic approximation of formal concept complexity
measures such as description-tree complexity \citep{distel2014concept}, which are
unsuitable here as they cover only EL rather than the full $\mathcal{SROIQ}$
expressivity of OWL~2.}

Concept expressions that differ minimally in NL are harder for LLMs to distinguish \citep{he-etal-2025-xcomps} and require formal reasoning, so we define a similarity metric over their
verbalisations, used for distractor ranking and class selection (Section~\ref{sec:overall}):
\begin{equation}
  \mathtt{sim}(C, C') \;=\;
  \frac{\mathbf{e}(\mathtt{smooth}(C))^\top\, \mathbf{e}(\mathtt{smooth}(C'))}
       {\|\mathbf{e}(\mathtt{smooth}(C))\|\;\|\mathbf{e}(\mathtt{smooth}(C'))\|}
  \label{eq:sim}
\end{equation}
where $\mathbf{e}(\cdot)$ denotes the text-embedding function and $\mathtt{smooth}(C)$
the smoothed natural-language verbalisation of $C$ from Stage~2.
For example, for the grain boundary axiom, the three highest-ranked distractors are
the child crystal swaps ($\mathtt{sim} = 0.976$, $0.966$, $0.954$), while the
cardinality contradiction ($\mathtt{sim} = 0.909$) and the existential weakening
($\mathtt{sim} = 0.856$) rank lower.

\subsection{Overall Pipeline}
\label{sec:overall}
Stage~1 produces two outputs from $\mathcal{O}$: a set $\mathcal{G}$ of class-indexed axiom groups each sorted by descending $\mathtt{oc}(C)$, and a named-class hierarchy index $\mathcal{H}$ built from inferred \texttt{subClassOf} triples.
Two filters are applied: \textsc{MinDepth} excludes axioms where $C$ is a single named class (trivially answerable), and \textsc{Primitives} excludes axioms whose fillers are entirely upper ontology classes too generic for domain-specific evaluation.
The remaining axioms form the working-set, which is extended with inferred axioms: a reasoner propagates the complex expressions of each entailed superclass to $A$ as $A \sqsubseteq C$, with duplicate $(A, C)$ pairs removed.
\textsc{MaxClassesPerAxiom} caps the number of classes sharing the same $C$ from a common ancestor, preventing a single popular ancestor expression from dominating the benchmark; when exceeded, a greedy max-min selection retains the most label-diverse subset, measured by Eq.~\ref{eq:sim} over class labels from $\mathcal{H}$.

Distractor candidates are generated for each axiom in the working-set, producing $\mathcal{W}_{\mathit{formal}} = \{(A,\, C,\, \mathit{distractor\_pool},\, \mathtt{oc}(C))\}$, where $\mathit{distractor\_pool} = \{(C',\, \mathit{category}(C'))\}$.
The disjoint index $\mathcal{D}$ is built by scanning $\mathcal{O}$ for disjointness axioms and expanding each entry to descendants via $\mathcal{H}$.
Operators are applied to each axiom in descending $\mathtt{oc}(C)$ order; when a class exceeds \textsc{MaxAxiomsPerClass}, only its most complex axioms are retained, ensuring representative coverage.
An axiom is included in $\mathcal{W}_{\mathit{formal}}$ only if its distractor pool contains at least \textsc{MinDistr} valid candidates; axioms below this threshold cannot yield complete MCQs.

In Stage~2, every concept expression in $\mathcal{W}_{\mathit{formal}}$ is verbalised
as described in Section~\ref{sec:verbalise}.

In Stage~3, each record is assigned its top-$\textsc{TopK}$ distractors, ranked first by \textsc{CategoryRank}, prioritising categories that require more precise formal reasoning (contradiction, weakening), then by descending $\mathtt{sim}(C, C')$ (Eq.~\ref{eq:sim}) within each category. Category rank thus controls the reasoning-type composition, while similarity maximises NL difficulty.
For the grain boundary axiom ($\textsc{TopK}$=3), the selected distractors are the cardinality contradiction ($\mathtt{sim} = 0.909$), the existential weakening ($\mathtt{sim} = 0.856$), and a child-filler non-entailed-satisfiable candidate ($\mathtt{sim} = 0.976$); the last ranks third despite its higher similarity because category rank takes precedence.

In Stage~4, each record is assembled into a complete MCQ and subjected to quality
filtering, producing $\mathcal{W}_{mcq}$.
Two stem variants are generated: a \emph{template stem} that fills a fixed pattern with the class label (e.g., ``Which of the following best describes grain boundary?'') and, since fixed templates cannot produce domain-contextualised prose, a \emph{scenario stem} generated by an LLM from the class label and domain description.
To prevent the stem wording from hinting at the correct answer, the LLM prompt includes a blocklist of non-stopword content tokens extracted from the smoothed verbalisations of $C$ and all $C'$, forbidden in the stem.
For the grain boundary record, the domain description is ``materials engineering''
and the blocklist is [\textit{exactly, member, crystal, angle, misalignment, relational, quality,
alloy, each}], producing the scenario stem: ``When evaluating the microstructure of
a polycrystalline metal, which of the following best describes a grain boundary?''.
The four options are assembled by randomly shuffling the correct answer and the
top-$\textsc{TopK}$ distractors, giving a random-guessing baseline of
$\frac{100}{\textsc{TopK}+1}$\% (25\% when $\textsc{TopK} = 3$). The complete assembled MCQ
for the grain boundary record is shown in Fig.~\ref{fig:workflow}.
Correctness with respect to the ontology's axiom set, and structural diversity, are guaranteed by design, but the NL quality of the assembled MCQ requires a separate check: following work showing that LLM judges are effective for filtering large MCQ benchmarks \citep{hashemi2024llm}, an LLM judge scores each item on three dimensions (integer 1--5): option faithfulness, stem clarity, and domain relevance.
The judge prompt is a fixed template that includes the MCQ together with the DL
expressions of the class definition axiom and the distractor expressions.
Records whose mean score meets or exceeds \textsc{QualityThresh} are retained in
$\mathcal{W}_{mcq}$.
The pipeline pseudocode is provided in the Appendix.

%% file: experiments.tex
\section{Experiments}
\label{sec:experiments}
Experiments validate three claims: 1) the pipeline produces MCQs grounded in the input ontology,
provided it is sufficiently axiomatised;
2) the distractors are formally incorrect with respect to the ontology
and test distinct forms of logical reasoning;
3) the MCQs meet the natural language quality standard required for a
meaningful benchmark.

\textbf{Setup.}
We consider three ontologies with contrasting scale and axiomatisation depth: 1) Pizza (85 classes, 77 disjointness declarations, 159 working-set axioms, $\mathtt{oc} \leq 2$, mean 1.28); 2) PMDco (1,071 classes, 15 disjointness declarations, 2,727 working-set axioms, $\mathtt{oc}(C)$ up to 6, mean 1.55, 299 axioms with $\mathtt{oc}(C) \geq 3$); and 3) DOID (9,969 classes, 26 disjointness declarations, 19,010 working-set axioms, $\mathtt{oc}(C)$ up to 17, mean 1.14, 679 axioms with $\mathtt{oc}(C) \geq 3$).

The pipeline uses $\textsc{TopK}{=}3$, $\textsc{MinDistr}{=}3$, $\textsc{QualityThresh}{=}3.0$, and $\textsc{Seed}{=}42$; distractors are ranked by $\textsc{CategoryRank} = [\textit{contradiction} \succ \textit{weakening} \succ \textit{non\_entailed\_satisfiable} \succ \textit{incomparable}]$. HermiT 1.3.8 accessed via Owlready2 \citep{lamy2017owlready} is used as the reasoner.
Stage~2 smoothing and Stage~4 stem generation use GPT-4.1 \citep{openai2025gpt41}, Stage~4 quality judging uses GPT-5.4 \citep{openai2025gpt5}, and Stage~3 similarity uses text-embedding-3-small \citep{openai_textembedding3small}.
Full implementation details including operator caps and prompts are provided in the Appendix.

\begin{figure}[!htbp]
    \centering   \includegraphics[width=1\linewidth]{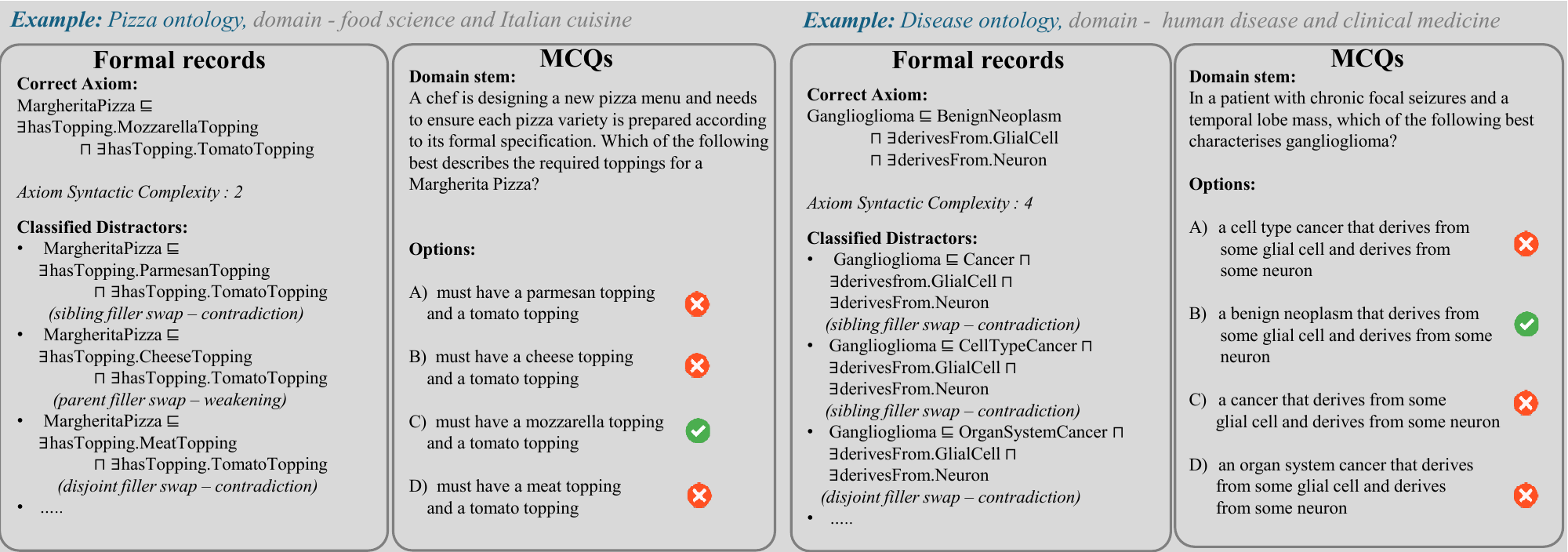}
    \vspace{-1.5em}
    \caption{ Example of MCQ generated using the proposed pipeline from Pizza (left) and DOID.}
    \label{fig:placeholder}
\end{figure}

\textbf{Pipeline Productivity.}
With this configuration, Pizza yields 112 MCQs; PMDco 2,491; and DOID 15,216 (Table~\ref{tab:pipeline_results}).
Example MCQs are in Fig.~\ref{fig:placeholder} (Pizza and DOID) and Fig.~\ref{fig:workflow} (PMDco), each showing the question, correct answer, DL grounding, and HermiT-verified distractor category.
Every MCQ is grounded in an ontology axiom, validating Claim~1.

\textbf{Formal Quality.}
Every correct answer is a class definition axiom; every distractor is HermiT-verified incorrect before entering the pool.
All four distractor categories are present in all three ontologies.
Pizza's 77 disjointness declarations produce a high contradiction rate (43.5\,\%); PMDco's sparse disjointness shifts the distribution toward non-entailed-satisfiable (61.4\,\%) and weakening (36.8\,\%). DOID's 26 disjointness declarations similarly suppress contradiction (1.9\,\%), with the distribution splitting between weakening (52.3\,\%) and non-entailed-satisfiable (45.7\,\%), reflecting its predominantly existential restriction axioms.
The distractor category distribution thus mirrors each ontology's original axiomatisation rather than a distribution imposed by the pipeline.
These results validate Claim~2.

\begin{table}[!htbp]
\vspace{-0.5em}
\scriptsize
\setlength{\tabcolsep}{3pt}\renewcommand{\arraystretch}{1.1}
\begin{minipage}[t]{0.62\linewidth}
\centering
\caption{Pipeline results across three ontologies: Pizza, PMDCo, and DOID.}
\label{tab:pipeline_results}
\begin{tabular}{lrrr@{\hspace{5pt}}lrrr}
\toprule
& \textbf{Pizza} & \textbf{PMDco} & \textbf{DOID}
& & \textbf{Pizza} & \textbf{PMDco} & \textbf{DOID} \\
\midrule
\multicolumn{4}{l}{\textbf{\# MCQs generated}}
& \multicolumn{4}{l}{\textbf{NL quality: sim.\ (0--1, $\uparrow$)}} \\
MCQ (\#)          & 112   & 2{,}491  & 15{,}216
& Avg.\ sim.       & 0.922 & 0.810 & 0.906 \\
\multicolumn{4}{l}{\textbf{Distractor categories (\%)}}
& \multicolumn{4}{l}{\textbf{NL quality: judge (1--5, $\uparrow$)}} \\
Contradiction     & 43.5  & 1.2   & 1.9
& Mean score       & 4.02  & 4.36  & 3.36 \\
Weakening         & 35.1  & 36.8  & 52.3
& Stem clarity     & 3.66  & 4.05  & 3.01 \\
Non-ent.\ sat.\   & 20.2  & 61.4  & 45.7
& Domain rel.\     & 4.08  & 4.23  & 2.27 \\
Incomparable      & 1.2   & 0.7   & 0.1
& Opt.\ faith.\    & 4.32  & 4.80  & 4.80 \\
\bottomrule
\end{tabular}
\end{minipage}
\hfill
\begin{minipage}[t]{0.35\linewidth}
\centering
\caption{LLM accuracy (\%).}
\label{tab:accuracy}
\begin{tabular}{lrrr}
\toprule
\textbf{Model label} & \textbf{Pizza} & \textbf{PMDco} & \textbf{DOID} \\
\midrule
Ministral-3-14B$^\circ$         & 46.4 & 45.7 & 54.1 \\
Qwen-3.8-27B$^\circ$            & 54.5   & 53.8  & 65.9   \\
Qwen-3.8-27B $^{\circ\dagger}$   & 62.5   & 59.5   & 68.4   \\
DeepSeek-R1$^{\circ\dagger}$    & 51.8   & 41.1   & 53.7   \\
GPT-5.6-sol$^{\bullet\dagger}$  & 76.8 & 74.9 & 75.3 \\
GPT-5.4$^\bullet$               & 69.6 & 64.2 & 66.1 \\
GPT-4o$^\bullet$                & 67.0 & 54.3 & 58.8 \\
\bottomrule
\end{tabular}

\vspace{2pt}
{\tiny $^\circ$~open\ $^\dagger$~reason\ $^\bullet$~frontier; random baseline: 25\%}
\end{minipage}
\end{table}
\textbf{NL Quality.}
Mean distractor similarity (Eq.~\ref{eq:sim}) is 0.922 (Pizza), 0.810 (PMDco), and 0.906 (DOID): distractors cannot be rejected on surface form alone.
Mean LLM judge score is 4.02 (Pizza), 4.36 (PMDco), and 3.36 (DOID) out of 5, all above the scale midpoint.
Lower Pizza scores reflect its simpler axiom structure, which yields less domain-specific stems; lower DOID scores reflect verbalisations that read unlike clinical prose, penalising domain relevance (2.27) despite near-perfect option faithfulness (4.80).
These results validate Claim~3.

As a preliminary check, a domain expert independently scored 20 PMDco items.
Expert mean (3.33) and LLM judge mean (4.0) both exceed the scale midpoint, with a mean absolute difference of 1.33, indicating that both raters rate quality similarly at the aggregate level and supporting the LLM judge as a practical quality filter; a larger human study with formal inter-rater agreement (e.g.\ Cohen's $\kappa$) may be investigated in future work.

Six LLMs (frontier, open-source, reasoning and non-reasoning) evaluated zero-shot (Table~\ref{tab:accuracy}) achieve 41.1--76.8\% accuracy, well above the 25\% random-guessing baseline, confirming that the benchmarks are challenging and discriminative (LLM details in Appendix, Table~\ref{tab:llm_details}).

%% file: conclusions.tex
\section{Conclusions}
Validating LLM reasoning over structured domain knowledge remains an open challenge for reliable scientific AI: manual benchmark construction does not scale, and LLM-generated benchmarks cannot be trusted to be logically sound.
This paper closes that gap with a pipeline that automatically turns any sufficiently axiomatised OWL 2 ontology into an MCQ benchmark, with every correct answer grounded in a class definition axiom and every distractor formally verified as incorrect by an OWL reasoner.
Applying the pipeline to three structurally contrasting ontologies (Pizza, PMDco, and DOID), we generated benchmarks of 112, 2,491, and 15,216 items with no manual annotation.
Practitioners building scientific AI systems can point the pipeline at their own domain ontology to obtain a diagnostic benchmark that reveals not just whether but where and how model reasoning breaks down, and to extract fine-tuning signal or neuro-symbolic supervision from the description logic formalisation and reasoner-verified labels.
Future work includes extending the distractor operators, strengthening the DeepOnto verbalisation templates, and quantifying the effect of verbalisation and smoothing on benchmark quality.
The most promising direction is extending benchmark generation to derived axioms (logical consequences entailed by the ontology but not explicitly asserted), which probe deeper reasoning and better isolate reasoning ability from prior knowledge of public ontologies.
Initial experiments on Pizza, reusing $\pi_{F_2}$ (ancestor filler substitution) to derive entailed axioms, already yield 191 additional MCQs.

%% file: appendix.tex
\section*{Appendix}
\subsection*{LLM Usage Disclosure}
LLMs are core components of the presented pipeline and the subjects of its evaluation; the models, roles, parameters, and prompt templates are described in Section~4 and this appendix. An LLM assistant was additionally used only for writing, editing, or formatting purposes and did not impact the core methodology, scientific rigorousness, or originality of the research.

\subsection*{Full Pipeline Implementation Details}
$\textsc{MaxClassesPerAxiom}{=}3$, $\textsc{MaxAxiomsPerClass}{=}3$; operator $\pi_F$ caps $N{=}2$, $P{=}4$, $M{=}6$ (Table~\ref{tab:operators}); seeded shuffle ($\textsc{Seed}{=}42$) applied when multiple candidates share the same similarity score.
BFO \citep{arp2015building} classes are used as primitives for PMDco; none for Pizza.
The reasoner is HermiT 1.3.8 (build 1.3.8.1099) via Owlready2 0.50 \citep{lamy2017owlready}, with a 180-second timeout per call; axioms exceeding this limit are skipped.
Stage~2 smoothing uses GPT-4.1 \citep{openai2025gpt41} (temperature~$= 0$, max tokens~$= 150$).
Stage~3 similarity uses text-embedding-3-small \citep{openai_textembedding3small} (1536 dimensions).
Stage~4 stem generation uses GPT-4.1 (temperature~$= 0.7$, max tokens~$= 120$); quality judging uses GPT-5.4 \citep{openai2025gpt5} (temperature~$= 0$, max tokens~$= 200$).
All models were selected based on availability during the period this research was conducted.
All GPT LLMs and text embedding models are accessed via Azure OpenAI \citep{azureopenai_api_2025_preview} with all other parameters set to their defaults as per the Azure OpenAI API.

\subsection*{Overall Pipeline Algorithm}
The overall pipeline with four stages is detailed in Algorithm~\ref{alg:overview}.
\begin{algorithm}[!h]
\footnotesize
\caption{Ontology-grounded MCQ Benchmark Generation (Overview)}
\label{alg:overview}
\begin{algorithmic}[1]
\Require[\textbf{User Inputs}] Ontology $\mathcal{O}$,\ domain description $D$
\Require[\textbf{Parameters}] $\textsc{MinDepth}$: minimum operator count $\mathtt{oc}(C)$,\
         $\textsc{MaxClassesPerAxiom}$: max classes $A$ retained per shared (ancestor, $C$) group,\
         $\textsc{MaxAxiomsPerClass}$: max axioms perturbed per class $A$,\
         $\textsc{MinDistr}$: minimum distractors required to retain an axiom,\
         $\textsc{TopK}$: number of distractors selected per axiom for the MCQ,\
         $\textsc{QualityThresh}$: minimum mean judge score to include a MCQ in the benchmark,\
         $\textsc{Seed}$: random seed for template assignment and option shuffling,\
         $\textsc{CategoryRank}$: distractor category preference order,\
         $\mathit{Primitives}$: set of upper-level ontology classes used to filter axioms,\
         $\textsc{Templates}$: set of stem templates,\
         $\textsc{SmoothPrompt}$: prompt template for smoothing,\
         $\textsc{StemPrompt}$: prompt template for stem scenario generation,\
         $\textsc{JudgePrompt}$: prompt template for MCQ quality scoring
\Require[\textbf{API calls/ Tools}] $\textsc{DeepOnto}$: rules-based verbaliser,\
         $\textsc{Embed}$: text embedding model,\
         $\textsc{LLM}$: LLM used for smoothing, stem generation, and judging
\Ensure $\mathcal{W}_{mcq}$: quality-filtered MCQ benchmark

\Statex \textbf{// Stage 1: Formal Record Generation}
\State $(\mathcal{G},\; \mathcal{H}) \gets \Call{SelectAxioms}{\mathcal{O},\ \textsc{MinDepth},\ \textsc{MaxClassesPerAxiom},\ \mathit{Primitives}}$
\State $\mathcal{W}_{formal} \gets \Call{PerturbAndClassify}{\mathcal{O},\, \mathcal{G},\, \mathcal{H},\, \textsc{MaxAxiomsPerClass},\, \textsc{MinDistr}}$
       \Comment{refer Operator Table~\ref{tab:operators}}

\Statex \textbf{// Stage 2: Verbalisation \& Smoothing}
\ForAll{$(A,\, C,\, \mathit{distractor\_pool},\, \cdot) \in \mathcal{W}_{formal}$}
    \ForAll{$e \in \{C\} \cup \{C' \mid (C', \cdot) \in \mathit{distractor\_pool}\}$}
        \State Verbalise $e$ w.r.t.\ $\mathcal{O}$ via \textsc{DeepOnto}; smooth via $\textsc{LLM}(\textsc{SmoothPrompt})$; store as $\mathrm{smooth}(e)$
    \EndFor
\EndFor

\Statex \textbf{// Stage 3: Distractor Ranking}
\ForAll{$(A,\, C,\, \mathit{distractor\_pool},\, \cdot) \in \mathcal{W}_{formal}$}
    \ForAll{$(C',\, \cdot) \in \mathit{distractor\_pool}$}
        \State $\mathrm{sim}(C, C')$ \Comment{Eq.~\ref{eq:sim}}
    \EndFor
    \State Sort $\mathit{distractor\_pool}$ by $\textsc{CategoryRank}(\mathit{category})$, then by $\mathrm{sim}(C, C') \downarrow$ within each category
    \State $\mathit{top\_k} \gets$ top $\textsc{TopK}$ candidates from sorted pool \Comment{$\downarrow$: descending order}
\EndFor

\Statex \textbf{// Stage 4: MCQ Assembly \& Quality Filtering}
\State $\mathcal{W}_{mcq} \gets \emptyset$
\ForAll{$(A,\, C,\, \mathit{top\_k},\, \cdot) \in \mathcal{W}_{formal}$}
    \State $\mathit{stem}_T \gets$ template for $\mathrm{label}(A,\mathcal{H})$ from \textsc{Templates} using \textsc{Seed}
    \State $\mathit{stem}_S \gets \textsc{LLM}\bigl(\textsc{StemPrompt}(\mathrm{label}(A,\mathcal{H}),\, D,\, \mathrm{blocklist}(C,\, \mathit{top\_k}))\bigr)$
           \Comment{blocklist: content words of $\mathrm{smooth}(C)$ and all $\mathrm{smooth}(C')$, forbidden in stem}
    \State $(\mathit{opts},\, \mathit{ans}) \gets \mathrm{shuffle}\bigl(\{C\} \cup \mathit{top\_k},\; \textsc{Seed}\bigr)$
    \State $\mathit{scores} \gets \textsc{LLM}\bigl(\textsc{JudgePrompt}(\mathit{stem}_S,\, \mathit{opts},\, \mathit{ans},  A, C, C')\bigr)$
    \If{$\mathrm{mean}(\mathit{scores}) \geq \textsc{QualityThresh}$}
        \State $\mathcal{W}_{mcq} \gets \mathcal{W}_{mcq} \cup \{(A,\, C,\, \mathit{top\_k},\, \mathit{stem}_T,\, \mathit{stem}_S,\, \mathit{opts},\, \mathit{ans})\}$
    \EndIf
\EndFor
\State \Return $\mathcal{W}_{mcq}$
\end{algorithmic}
\end{algorithm}

\subsection*{Concept Distractor Generation Algorithm}
The distractor generation algorithm of Stage~1 (Section~\ref{sec:distractorperturbation}) is provided in Algorithm~\ref{alg:distractor}.
\begin{algorithm}[!h]
\footnotesize
\caption{\textsc{PerturbAndClassify}: Concept Distractor Generation}
\label{alg:distractor}
\begin{algorithmic}[1]
\Require $\mathcal{O}$,\ $\mathcal{G}$,\ $\mathcal{H}$,\ \textsc{MaxAxiomsPerClass},\ \textsc{MinDistr}
\Require[\textbf{Tool}] \textsc{HermiT}: OWL reasoner
\Ensure $\mathcal{W}_{formal} = \{(A,\,C,\,\mathit{distractor\_pool},\,\mathtt{oc}(C))\}$

\Statex \textit{$\mathcal{D}$: disjoint index (owl:disjointWith + owl:AllDisjointClasses, expanded to descendants);\
operators $\pi \in \{\pi_{F_1},\pi_{F_2},\pi_{F_3},\pi_{F_4},\pi_{S_1},\pi_{Q_1},\pi_{Q_2},\pi_{Q_3}\}$: see Table~\ref{tab:operators};\
$\mathit{produced}$: axioms accepted so far for class $A$}

\State $\mathcal{D} \gets \mathrm{build\_disjoint\_index}(\mathcal{O},\,\mathcal{H})$
\State $\mathcal{W}_{formal} \gets \emptyset$
\ForAll{$(A,\,[(C_1, \mathtt{oc}(C_1)),(C_2, \mathtt{oc}(C_2)),\ldots]) \in \mathcal{G}$}
  \State $\mathit{produced} \gets 0$
  \ForAll{$C_i$ in order of $\mathtt{oc}\!\downarrow,$\quad \textbf{while} $\mathit{produced} < \textsc{MaxAxiomsPerClass}$}

    \Statex \hspace{2.2em}\textbf{// Generate candidates}
    \State $\mathit{candidates} \gets \{(C',\pi) \mid \pi \in \text{operators},\ C' \in \pi(C_i),\ C' \neq C_i\}$
           \Comment{duplicates removed}

    \Statex \hspace{2.2em}\textbf{// Phase 1: entailment + satisfiability (batched HermiT call)}
    \ForAll{$C' \in \mathit{candidates}$}
      \State $c_1 \gets [\mathcal{O} \models A \sqsubseteq C']$;\quad
             $\mathit{sat} \gets [\mathcal{O} \models A \sqcap C' \sqsubseteq \bot]$
    \EndFor

    \Statex \hspace{2.2em}\textbf{// Phase 2: subsumption between $C_i$ and $C'$ (batched, $c_1{=}\top$ only)}
    \ForAll{$C'$ with $c_1{=}\top$}
      \State $c_2 \gets [\mathcal{O} \models C_i \sqsubseteq C']$;\quad
             $c_3 \gets [\mathcal{O} \models C' \sqsubseteq C_i]$
    \EndFor

    \Statex \hspace{2.2em}\textbf{// Classify and filter}
    \State $\mathit{distractor\_pool} \gets \{\,(C',\,\mathit{category}(C'))\mid$ \Comment{refer Definition~\ref{eqn:category}}
    \Statex \hspace{4em}$\neg c_1 \wedge \mathit{sat}\ \Rightarrow\ \textit{contradiction}$
    \Statex \hspace{4em}$\neg c_1 \wedge \neg\mathit{sat}\ \Rightarrow\ \textit{non\_entailed\_satisfiable}$
    \Statex \hspace{4em}$c_1 \wedge c_2 \wedge \neg c_3\ \Rightarrow\ \textit{weakening}$
    \Statex \hspace{4em}$c_1 \wedge \neg c_2 \wedge \neg c_3\ \Rightarrow\ \textit{incomparable}\,\}$
           \Comment{equivalent/strengthening discarded}

    \If{$|\mathit{distractor\_pool}| \geq \textsc{MinDistr}$}
      \State $\mathcal{W}_{formal} \gets \mathcal{W}_{formal} \cup \{(A,\,C_i,\,\mathit{distractor\_pool},\,\mathtt{oc}(C_i))\}$
      \State $\mathit{produced} \gets \mathit{produced} + 1$
    \EndIf
  \EndFor
\EndFor
\State \Return $\mathcal{W}_{formal}$
\end{algorithmic}
\end{algorithm}
\newpage

\subsection*{Smoothing Prompt (Stage~2)}
The following template implements the Stage~2 smoothing described in Section~\ref{sec:verbalise}.

\noindent\textbf{System message}

\begin{lstlisting}
You are a natural language editor for ontology class descriptions.

You will receive two inputs:
  RAW:  a phrase produced by a rule-based verbaliser (may contain
        artifacts or an error string)
  EXPR: a label-substituted Description Logic expression giving
        the exact logical meaning

Your task: write a single fluent, grammatical English sentence
that accurately captures the meaning in EXPR. The sentence will
be used as an answer option in a multiple-choice question.

Rules:
- Keep all concept and property names exactly as they appear in EXPR.
- Conjuncts "C1 ⊓ C2 ⊓ ..." -> join with "and";
  disjuncts "C1 ⊔ C2 ⊔ ..." -> join with "or".
- "∃r.C"  -> "has r some C" or the most natural equivalent.
- "∀r.C"  -> "has r only C".
- ">=n r.C" -> "has at least n r C".
- "<=n r.C" -> "has at most n r C".
- "=n r.C"  -> "has exactly n r C".
- "¬C"      -> "is not a C".
- "∃r{v}"   -> "whose r is v".
- "∃r.Self" -> "that r itself".
- "⊤" -> "anything";  "⊥" -> "nothing".
- If RAW contains foreign-language labels or labels not present in
  EXPR, ignore RAW and use EXPR alone.
- If RAW is a "[verbalisation error: ...]" string, ignore it and
  use EXPR alone.
- Start the sentence with "something that" and end without trailing
  punctuation.
- Output ONLY the rewritten sentence -- no explanation, no label,
  no quotes.

--- Example 1: foreign-language labels in RAW -- ignore RAW ---
RAW:  EtwasdaseineTierOderPflanze
EXPR: ∃has part.animal ⊔ ∃has part.plant
-> something that has some animal or plant as a part

--- Example 2: chained relative clauses -- flatten with "and" ---
RAW:  something that has output document that is about legal entity
      that has role defendant role
EXPR: ∃has output.document ⊓ ∃is about.legal entity
      ⊓ ∃has role.defendant role
-> something that produces some document about a legal entity
   that has a defendant role

--- Example 3: verbalisation error -- ignore RAW, use EXPR ---
RAW:  [verbalisation error: member of -- ObjectExactCardinality(
      2 <hasRole>)]
EXPR: =2 has member.structural unit
-> something that has exactly 2 structural units as members
\end{lstlisting}

\noindent\textbf{User message}
\begin{lstlisting}
Rewrite the following into a fluent English sentence:
RAW:  <deeponto verbalisation or error string>
EXPR: <label-substituted DL expression>
\end{lstlisting}

\subsection*{Stem Generation Prompt (Stage~4)}
The following template generates the Stage~4 scenario stem (Section~\ref{sec:overall}).

\noindent\textbf{System message}
\begin{lstlisting}
You write multiple-choice question stems for professional knowledge
assessments.
Write as a subject-matter expert authoring an exam -- not as a
knowledge engineer.
Never use meta-language such as: ontology, ontological,
formalization, formal constraint, taxonomy, classification scheme,
or knowledge base. Write only in plain professional English.
\end{lstlisting}

\noindent\textbf{User message}
\begin{lstlisting}
Concept: <class label>
Domain: <domain string>
Definition: <SKOS definition, if available>

Write a single concise question stem (one sentence, max 30 words)
that:
- Names "<class label>" as the subject in a realistic <domain>
  professional context (e.g. classifying a material, evaluating a
  sample, designing a process, preparing a dish)
- Ends with "which of the following..." or "which of the following
  best..."
- Uses domain-professional language -- no meta-language about
  ontologies, definitions, or formalizations
- Does NOT mention any of these terms (they appear in the answer
  options and would hint at the answer): <blocklist>
- Does not describe or hint at the correct constraint

Output ONLY the stem -- no explanation, no quotes, no label.
\end{lstlisting}

\subsection*{Judge Prompt (Stage~4)}
The following template implements the Stage~4 LLM judge (Section~\ref{sec:overall}).

\noindent\textbf{System message}
\begin{lstlisting}
You are a rigorous examiner reviewing multiple-choice questions for
a professional knowledge assessment. Score each question on the three
dimensions provided.
Output only valid JSON -- no markdown fences, no explanation.
\end{lstlisting}

\noindent\textbf{User message}
\begin{lstlisting}
Domain: <domain string>
Class: <class label>

Question stem (scenario):
  <stem_scenario>

Options:
  A) <option text>
  B) <option text>
  C) <option text>
  D) <option text>

Correct answer: <letter>
Correct option text: <smoothed NL of correct expression>
Correct DL expression: <DL string>

Wrong options:
  <letter>) <option text>  [failure_type: <type>]
  ...

Scoring rubric (apply to each dimension):
  5 -- Excellent: fully meets the criterion with no issues
  4 -- Good: meets the criterion with minor issues
  3 -- Acceptable: partially meets the criterion, noticeable issues
  2 -- Poor: fails to meet the criterion in a significant way
  1 -- Unacceptable: completely fails the criterion

Dimension rubrics:
  option_faithfulness:
    5 -- NL text of correct option captures every constraint in the
         DL expression exactly, no omissions, no additions,
         no ambiguity
    4 -- Minor wording imprecision but meaning fully preserved
    3 -- One constraint is weakened, generalised, or slightly
         misleading
    2 -- A key constraint is missing or incorrectly stated
    1 -- NL text does not reflect the DL expression at all

  stem_clarity:
    5 -- Stem is a clear, natural professional question; no hint of
         the answer; domain-appropriate phrasing throughout
    4 -- Clear and appropriate with minor phrasing issues
    3 -- Understandable but awkward or slightly ambiguous
    2 -- Ambiguous, misleading, or contains a subtle hint toward
         the answer
    1 -- Confusing, domain-inappropriate, or directly hints at
         the answer

  relevance:
    5 -- A domain expert would find this question meaningful,
         appropriately challenging, and directly relevant to
         professional practice
    4 -- Relevant and meaningful with minor reservations
    3 -- Somewhat relevant but too easy, too obscure, or contrived
    2 -- Marginally relevant; most domain experts would find it
         trivial or odd
    1 -- Not relevant to the domain or incomprehensible to a
         professional

Score the question on all three dimensions and output ONLY valid JSON:
{
  "option_faithfulness": <1-5>,
  "stem_clarity": <1-5>,
  "relevance": <1-5>,
  "notes": "<one sentence on main quality issue, or 'none'>"
}
\end{lstlisting}

\subsection*{Benchmark Schema}
The complete schema of each benchmark record is illustrated in Table~\ref{tab:schema}.
\begin{table}[!htbp]
\centering
\small
\caption{Complete Benchmark Record Schema}
\label{tab:schema}
{\setlength{\tabcolsep}{5pt}\renewcommand{\arraystretch}{0.88}
\begin{tabular}{>{\footnotesize}l >{\footnotesize}l >{\footnotesize}p{6.5cm}}
\toprule
\footnotesize\textbf{Field} & \footnotesize\textbf{Type} & \footnotesize\textbf{Description} \\
\midrule

\multicolumn{3}{l}{\textit{Class identity}} \\
\texttt{class\_iri}            & string & Ontology IRI of subject class\\
\texttt{class\_label}          & string & rdfs:label of class\\

\texttt{inferred}              & bool   & True if axiom inherited via HermiT\\
\texttt{axiom complexity}       & integer & $\mathtt{oc}$ (Section~\ref{sec:metric})\\

\midrule

\multicolumn{3}{l}{\textit{Correct expression}} \\
\texttt{correct\_expression}   & string & Owlready2 string of $C$\\

\texttt{correct.dl}            & dict & Recursive JSON encoding of $C$\\
\texttt{correct.smooth}        & string & Fluent English option string\\
\midrule

\multicolumn{3}{l}{\textit{Distractor pool (per distractor entry)}} \\

\texttt{distractors[i].failure\_type}   & enum  &
        contradiction $|$ weakening $|$ incomparable $|$
        non\_entailed\_satisfiable \\

\texttt{distractors[i].similarity}      & float  & similarity as per Eq.~\ref{eq:sim} to correct option\\
\texttt{distractors[i].rank}            & int    & Rank as per Stage 3\\
\texttt{distractors[i].dl}              & dict &  $C'$ in JSON encoding\\
\texttt{distractors[i].smooth}          & string & Fluent English option string\\
\texttt{distractor\_pool\_size}         & int   & Total candidates before filter\\

\midrule

\multicolumn{3}{l}{\textit{Top-$k$ distractors (selected for MCQ)}} \\
\texttt{top\_k}                & list   & Top $k{=}3$ distractors\\
\midrule

\multicolumn{3}{l}{\textit{Question stems}} \\
\texttt{stem\_template}        & string & Template stem, seeded by class IRI\\
\texttt{stem\_scenario}        & string & LLM domain scenario stem\\
\midrule

\multicolumn{3}{l}{\textit{MCQ options}} \\
\texttt{options}               & dict   & \{A,B,C,D\} $\rightarrow$ option string;
                                          reproducible seeded shuffle\\
\texttt{correct\_answer}       & A--D   & Letter of correct option\\
\midrule

\multicolumn{3}{l}{\textit{Quality scores}} \\

\texttt{quality\_mean}         & float 1--5 & Mean of three dimension scores\\
\texttt{quality\_flag}         & bool   & \textsc{True} if mean $\geq 3.0$\\
\texttt{quality\_notes}        & string & One-sentence judge summary\\
\bottomrule
\end{tabular}
}
\end{table}

\subsection*{LLM Evaluation Setup}
We evaluate six models spanning proprietary and open-source, reasoning and non-reasoning varieties, as detailed in Table~\ref{tab:llm_details}.
Proprietary models (GPT-4o \citep{openai2024gpt4o}, GPT-5.4 \citep{openai2025gpt5}, GPT-5.6-sol \citep{openai2025gpt56}) are accessed via the Azure OpenAI API \citep{azureopenai_api_2025_preview} with all parameters set to their defaults except temperature (0) and maximum output tokens (4,096).
Open-source models (Ministral-3-14B \citep{mistral2025ministral}, Qwen-3.8-27B \citep{qwen2025qwen3827b}, DeepSeek-R1 \citep{deepseek2025r1qwen32b}) are loaded locally via HuggingFace Transformers \citep{wolf-etal-2020-transformers} and served with the same inference settings.
Qwen-3.8-27B is evaluated in two configurations: \textit{thinking off} (standard generation, non-reasoning) and \textit{thinking on} (extended chain-of-thought enabled via the model's native thinking flag), allowing direct comparison of the effect of the reasoning mode on the same base model.

\begin{table}[!htbp]
\centering
\caption{LLMs used for zero-shot MCQ evaluation. Prop.\ = proprietary (Azure OpenAI); Open = open-source (HuggingFace); HF = HuggingFace; R = reasoning mode enabled; NR = non-reasoning.}
\label{tab:llm_details}
\small
\begin{tabular}{llll}
\toprule
\begin{tabular}[c]{@{}l@{}}\textbf{Model label}\\\textbf{(Table~\ref{tab:accuracy})}\end{tabular} & \textbf{Model / HF ID} & \textbf{Access} & \textbf{Mode} \\
\midrule
GPT-5.6-sol     & \texttt{gpt-5.6-sol} (Azure OpenAI) & Prop. & R \\
GPT-5.4         & \texttt{gpt-5.4} (Azure OpenAI)     & Prop. & NR \\
GPT-4o          & \texttt{gpt-4o} (Azure OpenAI)      & Prop. & NR \\
Ministral-3-14B & \texttt{mistralai/Ministral-3-14B-Instruct-2512} (HF) & Open & NR \\
Qwen-3.8-27B    & \texttt{Qwen/Qwen3.8-27B} thinking off (HF) & Open & NR \\
Qwen-3.8-27B    & \texttt{Qwen/Qwen3.8-27B} thinking on (HF)  & Open & R  \\
DeepSeek-R1-Distill-32B     & \texttt{deepseek-ai/DeepSeek-R1-Distill-Qwen-32B} (HF) & Open & R \\
\bottomrule
\end{tabular}
\end{table}

All models are evaluated zero-shot with temperature 0, a maximum of 4,096 output tokens, a domain-specific system prompt, and no chain-of-thought, ensuring purely direct letter prediction.

All models are prompted with the following zero-shot template:

\noindent\textbf{System message}
\begin{lstlisting}
You are a [domain expert]. Answer the multiple-choice question
below by replying with ONLY the single letter of the correct
option (A, B, C, or D). Do not add any explanation.
\end{lstlisting}

\noindent\textbf{User message}
\begin{lstlisting}
<stem_scenario>

A. <option_A>
B. <option_B>
C. <option_C>
D. <option_D>

Answer (A/B/C/D):
\end{lstlisting}
The [Domain expert] phrase varies per ontology to provide the domain specific instructions of the task: \textit{an expert in food science and Italian cuisine} (Pizza), \textit{an expert in process metallurgy and materials science} (PMDco), \textit{a medical expert in human disease and clinical medicine} (DOID).